\pdfoutput=1
\documentclass[runningheads]{llncs}
\usepackage{times}
\usepackage{helvet}
\usepackage{courier}
\usepackage{graphicx}
\usepackage{epstopdf}
\usepackage{latexsym}
\usepackage{amssymb}
\usepackage{amsmath}
\usepackage{enumitem}
\usepackage{framed}
\usepackage[colorinlistoftodos, textwidth=3cm]{todonotes} 
\usepackage[spaces,hyphens]{url}
\usepackage{xspace}
\usepackage{array}
\usepackage{listings}
\usepackage{hyperref}

\lstdefinelanguage{ttl}{
sensitive=true,
morekeywords={spr,bind,sequenceOf,evalOf,fixPointOf,Rule,head,body},
keywordstyle=\bf\ttfamily,
tabsize=2,
basicstyle=\fontsize{8.5}{8.5}\selectfont\ttfamily,
showstringspaces=false
}

\title{Knowledge Propagation in Contextualized Knowledge Repositories:
an Experimental Evaluation}

\author{Loris Bozzato\and Luciano Serafini}

\titlerunning{Knowledge Propagation in CKR: an Experimental Evaluation}

\institute{
  Fondazione Bruno Kessler,\\ 
  Via Sommarive 18, 38123 Trento, Italy\\
 \email{\texttt{\{bozzato,serafini\}@fbk.eu}}
}

\def\C{\mathcal{C}}	% context
\def\CKB{\mathfrak{K}}

\def\I{\mathcal{I}}

\newcommand{\Pair}[2]{\left\langle#1,#2\right\rangle}

\newcommand{\SROIQ}{\mathcal{SROIQ}}

\newcommand{\IC}{\mathfrak{I}}

\newcommand{\R}{\mathcal{R}}
\newcommand{\A}{\mathcal{A}}

\newcommand{\stru}[1]{\langle #1 \rangle}

\newcommand{\non}{\neg}

\newcommand{\subs}{\sqsubseteq}

\newcommand{\Lcal}{{\cal L}}

\newcommand{\Mcal}{{\cal M}}

\newcommand{\cov}[1]{\preceq}

\newcommand{\nop}[1]{}

\setitemize{label=--,leftmargin=*}
\setenumerate{leftmargin=*}
\newcommand{\gray}[1]{\textcolor{gray}{#1}}

\renewcommand{\-}{\text{-}}

\newcommand{\SROIQrl}{\mathcal{SROIQ}\text{-RL}}

\newcommand{\KB}{\mathrm{K}} %knowledge base of module/context

\newcommand{\mlc}{\ml{c}}
\newcommand{\mlm}{\ml{m}}

\newcommand{\mlmod}{\ml{mod}}
\newcommand{\Ctx}{\ml{Ctx}}

\newcommand{\Gl}{\mathfrak{G}}

\newcommand{\N}{\boldsymbol{\mathsf{N}}}
\newcommand{\M}{\boldsymbol{\mathsf{M}}}
\renewcommand{\C}{\boldsymbol{\mathsf{C}}}
\renewcommand{\R}{\boldsymbol{\mathsf{R}}}
\renewcommand{\A}{\boldsymbol{\mathsf{A}}}
\newcommand{\ml}[1]{\mathsf{#1}} %meta-language object

\newcommand{\subClass}{{\tt subClass}}

\newcommand{\triple}{{\tt triple}}

\newcommand{\subEval}{{\tt subEval}}

\newcommand{\eval}{\textsl{eval}}

\begin{document}

\maketitle
%\vspace*{-1\baselineskip}
%%
%% Proceedings Production
%%
\setcounter{page}{13}
\thispagestyle{plain}

\begin{abstract}
  % # Intro #
	As the interest in the representation of context dependent knowledge in the Semantic Web has
  been recognized, a number of logic based solutions have been proposed in this regard. 
	In our recent works, in response to this need, we presented the description
  logic-based Contextualized Knowledge Repository (CKR) framework. 
  %
  % # The CKR over RDF #
	CKR is not only a theoretical framework, but it has been effectively implemented 
	over state-of-the-art tools for the management of Semantic Web data:
	%we recently presented an implementation for the CKR in which 
	inference inside and across contexts has been realized in the form of 
	forward SPARQL-based rules over different RDF named graphs.
  %
  % # In this paper we contribute with...	#
	In this paper we present the first evaluation results for such
	CKR implementation. %of the CKR framework.
	In particular, in first experiment we study its \emph{scalability} with respect to 
	different reasoning regimes. %and storage solutions.
	In a second experiment we analyze the effects of \emph{knowledge propagation}
	on the computation of inferences.
\end{abstract}

%-----------------------------------------------------------------------------

%  \input{sec/intro}

%-----------------------------------------------------------------------------
\section{Introduction}
\label{sec:intro}

% # Intro: contexts in the SW #
Recently, the representation of context dependent knowledge in the Semantic Web has
been recognized as a relevant issue. 
This lead to the introduction of a growing number of
logic based proposals, e.g.\
\cite{CDF,klarman:13,serafini-homola-ckr-jws-2012,stra-etal-2010,Tanca:07,udrea-annotated-rdf-2010}.
In this line of research, in our previous works we introduced the
Contextualized Knowledge Repository (CKR) framework
\cite{serafini-homola-ckr-jws-2012,BozzatoSerafini:13,BozGhiSer:KCAP2013,BozHomSer:DL2012}.
CKR is a description logics-based framework defined as a two-layered structure: 
intuitively, a lower layer contains a set of contextualized knowledge bases, while 
the upper layer contains context independent knowledge and meta-data defining the structure
of contextual knowledge bases.

% # CKR framework implemented over RDF #
The CKR framework has not only been presented as a theoretical framework,
but we also proposed effective implementations based on its definitions~\cite{BozzatoSerafini:13,BozzatoEiterSerafini:14}.
In particular, in~\cite{BozzatoSerafini:13} we presented an implementation for the CKR framework
over state-of-the-art tools for storage and inference over RDF data.
%\lbnote{CITARE ESEMPI DI APPLICAZIONE DEL CKR?}
Intuitively, the CKR architecture can be implemented by representing the global context and the
local object contexts as distinct RDF named graphs.
Inference inside (and across) named graphs is implemented as
SPARQL based forward rules. We use an extension of
the Sesame framework that we developed, called \emph{SPRINGLES}, which provides methods to demand an inference
materialization over multiple graphs: rules are encoded as SPARQL queries and it is possible to customize
their evaluation strategy. The rule set encodes the rules of the formal materialization calculus 
we proposed for the CKR framework~\cite{BozzatoSerafini:13} and the evaluation strategy follows the calculus 
translation process.

% # In this paper we present an evaluation #
In this paper we present the results of an initial experimental evaluation of such implementation of 
CKR framework over RDF.
%
% # Research questions and findings #
In particular, the experiments we present are aimed at answering two different research questions:
\begin{itemize}
\item 
  \textbf{RQ1 (scalability):} \emph{what is the effect on the amount of time requested for
  inference closure computation with respect to the number and size of contexts of a CKR?}
  %(w.r.t. different solutions for inference regimes)
\item 
  \textbf{RQ2 (propagation):} \emph{what is the effect on the amount of time requested for
  inference closure computation with respect to the number of connections across contexts?}
  (considering a fixed number of contexts and a fixed amount of knowledge exchanged).
\end{itemize}
As we will detail in the following sections,
by means of our experiments we answered the questions with these findings:
\begin{itemize}
\item 
  \textbf{F1 (scalability):}
  reasoning regime at the global and local level strongly impacts on the 
  scalability of reasoning and its behaviour. Considering only 
  global level reasoning, results suggest that the management of contexts
  does not add overhead to the reasoning in global context; 
  by considering also reasoning inside contexts, inference time 
  appears to be influenced by the expressivity and number of contexts.
\item 
  \textbf{F2 (propagation):}
  knowledge propagation cost linearly depends on the number of connections.
  %the average increase in closure time for $k$ local connections (w.r.t. 
  %having no connections) amounts to $(5,\!1 \cdot k)\%$.
  Moreover, the representation of references to local interpretation of symbols 
	using %the CKR model that makes use of 
	context connections is always more compact
  w.r.t. replicating symbols for each local interpretation:
  the first solution in general requires more computational time,
  but outperforms the second solution in case of a larger number of connections.
\end{itemize}
%
% # Blueprint/Contributions #
The remainder of the paper is organized as follows:
%in Section 2 we provide some preliminary definitions
in Section~\ref{sec:ckr} we summarize the basic formal definitions for CKR and its associated calculus;
in Section~\ref{sec:springles} we summarize how the presented definitions have been implemented over RDF named graphs;
in Section~\ref{sec:experiments} we present the test setup and experimental evaluations;
finally, in Section~\ref{sec:conclusion} 
we suggest some possible extensions to the current evaluation and implementation work.

%\lbnote{distribuire prototipo e testsets?}

%##########################

  %\input{sec/prelims}
  %\input{sec/CKR}

%-----------------------------------------------------------------------------
\section{Contextualized Knowledge Repositories}
\label{sec:ckr}

%<v. DL2014>
%<v. DL2014 slides>

\renewcommand{\C}{\boldsymbol{\mathsf{C}}}
\renewcommand{\R}{\boldsymbol{\mathsf{R}}}
\renewcommand{\A}{\boldsymbol{\mathsf{A}}}

\renewcommand{\inst}{{\tt inst}}

In the following we provide an informal summary of the definitions for the CKR framework: 
for a formal and detailed description and for complete examples, we refer
to~\cite{BozzatoSerafini:13} where the current formalization for CKR has been first introduced.

% # Structure of a CKR #
Intuitively, a CKR is a two layered structure:
%\lbnote{FIGURA?}
the upper layer consists of a knowledge base $\Gl$ containing (1) \emph{meta-knowledge}, i.e. the structure
and properties of contexts of the CKR, and (2) \emph{global (context-independent) knowledge},
i.e., knowledge that applies to every context; the lower layer consists of a set of (local)
contexts that contain (locally valid) facts and can refer to what holds in other contexts.
%The upper layer is composed of a
%knowledge base $\Gl$, which describes two types of knowledge: 
%(i) the structure and the properties of contexts of the CKR 
%(called \emph{meta-knowledge}), and 
%(ii) the knowledge that is context independent, i.e., that holds in every context 
%(called \emph{global knowledge}).  
%The lower layer, is constituted by a set of (local)
%contexts, each containing (locally valid) facts, that can also refer
%to what holds in other contexts. 
%To favor knowledge reuse, the knowledge of each context is
%organized in multiple knowledge modules. 
%The association between
%contexts and modules is represented in the meta-knowledge via a binary
%relation, and can be either explicitly asserted or inferred via
%meta-reasoning. 

%- - - - - - - - - - - - - - - - - - - - - - - - - - - - - - - - - - - - - - -
\smallskip\noindent
\textbf{Syntax.}
% # Meta-vocabulary #
%[Difference between meta-vocabulary and object vocabulary]
In order to separate the elements of the meta-knowledge
from the ones of the object knowledge, we build CKRs over
two distinct %(but not necessarily disjoint) 
vocabularies and languages.
The meta-knowledge of a CKR is expressed
in a DL language containing the elements that define the contextual structure.
A \emph{meta-vocabulary} is a DL vocabulary $\Gamma$
 containing the sets of symbols for 
\emph{context names} $\N$;
\emph{module names} $\M$;
\emph{context classes} $\C$, including the class $\ml{Ctx}$;
\emph{contextual relations} $\R$;
\emph{contextual attributes} $\A$;
and for every attribute $\ml{A} \in \A$, a set $\ml{D_A}$
	  of \emph{attribute values} of $\ml{A}$.
%\lbreview{aggiungere esempio di ognuno?}		
% # Mod relation #
The role $\mlmod$ defined on $\N \times \M$ expresses associations
between contexts and modules.
Intuitively, modules represent pieces of knowledge specific to 
a context or context class; attributes describe contextual dimensions 
(e.g. time, location, topic) identifying a context (class).
The \emph{meta-language} $\Lcal_\Gamma$ of a CKR
is a DL language over $\Gamma$ (where, formally, the 
range and domain of attributes and $\mlmod$ are restricted as explained above).
%such that
%in every concept $\bullet \ml{A.B}$ and $\bullet \mlmod.\ml{B}$ 
%where 
%$\bullet \in \{\forall, \exists, \leqslant n,\geqslant n\}$,
%concept $\ml{B}$ has the form $\ml{B}=\{\ml{a}\}$ with $\ml{a}\in
%\ml{D_A}$ respectively $\ml{B}=\{\ml{m}\}$ with $\ml{m}\in\M$. 

% # Object vocabulary #
The knowledge in contexts of a CKR is expressed via a DL
language $\Lcal_\Sigma$, called 
\emph{object-language}, based on an object-vocabulary $\Sigma$.  
%The object-language $\Lcal_\Sigma$ is the DL language
%defined starting from the $\Sigma$.
The expressions of the object language are evaluated locally to
each context, i.e., contexts can interpret each symbol
independently. 
To access the interpretation of expressions inside a specific context or context class, 
we extend $\Lcal_\Sigma$ to $\Lcal^e_\Sigma$
with \emph{eval expressions} 
of the form $\eval(X,\ml{C})$, where $X$ is a concept or role expression 
of $\Lcal_\Sigma$ and $\ml{C}$ is a concept expression of $\Lcal_\Gamma$
(with $\ml{C} \subs \Ctx$).
%\lbreview{aggiungere esempio di eval?}
Intuitively, $\eval(X,\ml{C})$ can be read as ``the interpretation of $X$
in all the contexts of type $\ml{C}$''.

% # CKR #
On the base of previous languages, %for the meta- and object-languages,
we define a \emph{Contextualized Knowledge Repository (CKR)}
as a structure $\CKB = \stru{\Gl, \{\KB_\mlm\}_{\mlm \in \M}}$ where:
(i) $\Gl$ is a DL knowledge base over $\Lcal_\Gamma\cup\Lcal_\Sigma$; 
(ii) every $\KB_\mlm$ is a DL knowledge base over $\Lcal^e_\Sigma$, for each module name $\ml{m} \in \M$. 
%
% # SROIQrl #
The knowledge in a CKR can be expressed by means of
any DL language: %, for every DL language we have a different CKR. 
%in the following we provide the parametric definition to any
%DL language and successively we instantiate it to $\SROIQrl$.
in this paper, we consider $\SROIQrl$ (defined in~\cite{BozzatoSerafini:13}) as language of reference.
$\SROIQrl$ is a restriction of $\SROIQ$ syntax corresponding to OWL RL~\cite{Motik:09:OWO}.
$\CKB$ is a $\SROIQrl$ CKR, if $\Gl$ and all $K_{\ml{m}}$ 
%are $\SROIQrl$ knowledge base,
are knowledge bases over the extended language of $\SROIQrl$
where eval-expressions can only 
occur in left-concepts and contain left-concepts or roles. 

%- - - - - - - - - - - - - - - - - - - - - - - - - - - - - - - - - - - - - - -
\smallskip\noindent
\textbf{Semantics.}
The model-based semantics of CKR basically follows the 
two layered structure of the framework.
A \emph{CKR interpretation} is a structure $\IC = \stru{\Mcal, \I}$ s.t.:
(i) $\Mcal$ is a DL interpretation of $\Gamma \cup \Sigma$ 
(respecting the intuitive interpretation of $\Ctx$ as the class of all contexts);
%s.t., %$\N^\Mcal \subseteq \ml{Ctx}^\Mcal$ 
%for every $\ml{c} \in \N$, $\ml{c}^\Mcal \in \ml{Ctx}^\Mcal$
%and, for every $\ml{C} \in \C$, $\ml{C}^\Mcal \subseteq \ml{Ctx}^\Mcal$;
(ii) for every $x\in\ml{Ctx}^\Mcal$, 
$\I(x)$ is a DL interpretation over $\Sigma$ 
(with same domain and interpretation of individual names of $\Mcal$).
%s.t.		$\Delta^{\I(x)} = \Delta^{\Mcal}$ and,
%for $a \in \NIs$ and any $y \in \Ctx^\Mcal$, $a^{\I(x)} = a^{\I(y)} = a^{\Mcal}$.
The interpretation of ordinary DL expressions on $\Mcal$ and $\I(x)$ 
in $\IC = \stru{\Mcal, \I}$ is as usual;
$\eval$ expressions are interpreted as follows: for every $x \in \Ctx^\Mcal$,
%\begin{center}\footnotesize
 $\eval(X,\ml{C})^{\I(x)} = \bigcup_{\ml{e} \in \ml{C}^{\Mcal}} X^{\I(\ml{e})}$, 
%\end{center}
i.e. the union of all elements in $X^{\I(\ml{e})}$ for all contexts $e$ in $\ml{C}^{\Mcal}$.

%According to the previous definition, a CKR interpretation is composed
%by an interpretation for the ``upper-layer'' (which includes the global
%knowledge and the meta-knowledge) and 
%an interpretation of the object language for each instance of type
%context (i.e., for all $x\in \ml{Ctx}^\Mcal$), 
%providing a semantics of the object-vocabulary in $x$.

% # CKR model #
A CKR interpretation $\IC$ is a \emph{CKR model} of $\CKB$ %(in symbols, $\IC\models\CKB$) 
iff the following conditions hold:
(i)   for $\alpha \in \Lcal_\Sigma \cup \Lcal_\Gamma$ in $\Gl$, $\Mcal \models \alpha$;
(ii)  for $\Pair{x}{y} \in \ml{mod}^\Mcal$ with $y= \ml{m}^\Mcal$, $\I(x)\models \KB_{\ml{m}}$;
(iii) for $\alpha \in \Gl \cap \Lcal_\Sigma$ and $x \in \Ctx^\Mcal$,
     $\I(x) \models \alpha$.
Intuitively, while the first two conditions impose that $\IC$
verifies the contents of global and local modules associated to contexts,
last condition states that global knowledge has to be propagated to local contexts.

%% # Reasoning tasks #
%%Similarly to~\cite{serafini-homola-ckr-jws-2012},
%We adapt to CKR models the definition of 
%the classic reasoning problem of entailment:
%intuitively, the problem is specialized by indicating the context of reference.
%
%\begin{definition}[$\mlc$-entailment]
%Given a CKR $\CKB$ over $\Pair{\Gamma}{\Sigma}$ with $\mlc\in\N$
%and an axiom $\alpha \in \Lcal^e_\Sigma$, 
%we say that $\alpha$ is $\mlc$-entailed by $\CKB$ 
%(denoted by $\CKB\models \mlc:\alpha$) if, for every CKR model
%$\IC = \stru{\Mcal, \I}$ of $\CKB$, 
%we have $\I(\mlc^\Mcal) \models \alpha$.
%\end{definition}
%%
%We extend this definition to CKR knowledge bases as follows.
%%
%\begin{definition}[Global entailment]
%  Given a CKR $\CKB$ over $\Pair{\Gamma}{\Sigma}$
%  and an axiom $\alpha$, 
%  we say that $\alpha$ is (globally) entailed by $\CKB$ 
%  (denoted by $\CKB\models \alpha$) if:\\[-4ex]
%  \begin{itemize}
%	\item 
%	  $\alpha \in \Lcal^e_\Sigma$ and, 
%	  for every $\mlc \in \N$ and 
%    model $\IC = \stru{\Mcal, \I}$ of $\CKB$, 
%%    we have $\I(\mlc),[\this/\mlc]\!\models\!\alpha$;
%    we have $\I(\mlc^\Mcal)\!\models\!\alpha$;
%  \item
%    $\alpha \in \Lcal_\Gamma$ and, 
%    for every 
%    model $\IC = \stru{\Mcal, \I}$ of $\CKB$,
%    we have $\Mcal \models \alpha$.
%  \end{itemize}
%\end{definition}

%- - - - - - - - - - - - - - - - - - - - - - - - - - - - - - - - - - - - - - -
\smallskip\noindent
\textbf{Materialization calculus.}
Reasoning inside a CKR has been formalized in form of 
a materialization calculus. 
In particular, the calculus proposed in~\cite{BozzatoSerafini:13}
is an adaptation of the calculus presented in~\cite{Krotzsch:10}
in order to define a reasoning procedure for deciding instance checking 
in the structure of a $\SROIQrl$ CKR. 
%
%In our presentation, we consider $\SROIQrl$ UKBs in normal form
%that include all of the previously presented metatheories.
As we discuss in following sections,
this calculus provides the formalization for the definition of rules for the 
implementation of CKR based on RDF named graphs and forward SPARQL rules.

%- - - - - - - - - - - - - - - - - - - - - - - - - - - - - - - - - - - - - - -
%\subsection{Calculus definition}

%We instantiate and adapt the general definition of \emph{materialization calculus}
%given by~\cite{Krotzsch:10} 
%in order to meet the two-layered structure of CKR and the language of $\SROIQrl$.

Intuitively, the calculus is based on a translation to datalog: 
the axioms of the input CKR are translated to datalog atoms and 
datalog rules are added to such translation to encode the global and local inferences
rules; instance checking is then performed by translating the ABox assertion to
be verified as a datalog fact and verifying whether it is entailed by the CKR program.
The calculus, thus, has three components:
(1) the \emph{input translations} $I_{glob}$, $I_{loc}$, $I_{rl}$, 
where given an axiom $\alpha$ and $\mlc \in \N$, each $I(\alpha, \mlc)$
is a set of datalog facts or rules: 
intuitively, they encode as datalog facts the contents of input global and local DL knowledge bases; 
(2) the \emph{deduction rules} $P_{loc}$, $P_{rl}$, 
which are sets of datalog rules: they represent the inference rules for the 
instance-level reasoning over the translated axioms;
and (3) the \emph{output translation} $O$, where
given an axiom $\alpha$ and $\mlc \in \N$, $O(\alpha, \mlc)$ is
a single datalog fact encoding the ABox assertion $\alpha$ that we want to prove 
 to be entailed by the input CKR (in the context $\mlc$).

We briefly present here the form of the different sets of 
translation and deduction rules:
tables with the complete set of rules can be found in~\cite{BozzatoSerafini:13}.

\noindent (i) {\em $\SROIQrl$ translation}: 
Rules in $I_{rl}(S, c)$ translate to datalog facts
$\SROIQrl$ axioms (in context $c$). E.g., we translate
atomic concept inclusions with the rule $A \subs B \mapsto \{\subClass(A,B,c)\}$.
The rules in $P_{rl}$ are the deduction rules
corresponding to axioms in $\SROIQrl$: e.g., for atomic concept inclusions
we have

\smallskip
\centerline{$\subClass(y,z,c), \inst(x,y,c) \to \inst(x,z,c)$}
\smallskip

\noindent (ii) {\em Global and local translations}: 
Global input rules of $I_{glob}$
encode the interpretation of $\Ctx$ in the global context.
Similarly, local input rules $I_{loc}$ and local deduction rules $P_{loc}$
provide the translation and rules for elements of the local object language. In particular for
$\eval$ expressions in concept inclusions, 
we have the input rule $\eval(A, \ml{C}) \subs B \mapsto \{\subEval(A, \ml{C}, B, \mlc)\}$
and the corresponding deduction rule (where $\ml{g}$ identifies the global context):

\smallskip
\centerline{$\subEval(a, c_1, b, c), \inst(c', c_1, \ml{g}), \inst(x, a, c') \to \inst(x, b, c)$}
\smallskip

\noindent(iii) {\em Output rules}: The rules in $O(\alpha, \mlc)$
provide the translation of ABox assertions that can be verified to hold in context $c$ by applying the rules of the final program. For example, atomic concept assertions in a context $\mlc$ are 
translated by $A(a) \mapsto \{\inst(a,A,\mlc)\}$.

%- - - - - - - - - - - - - - - - - - - - - - - - - - - - - - - - - - - - - - -
%\subsection{Translation process}

Given a CKR $\CKB = \stru{\Gl, \{\KB_\mlm\}_{\mlm \in \M}}$,
the translation to its datalog program $PK(\CKB)$ proceeds in four
steps:
\begin{enumerate}
\item 
  the \emph{global program} $PG(\Gl)$ for $\Gl$ is translated by applying input rules
  $I_{glob}$ and $I_{rl}$ to $\Gl$ and adding deduction rules $P_{rl}$;
\item
  Let $\N_\Gl = \{\mlc \in \N \;|\; PG(\Gl) \models \inst(\mlc,\Ctx,\ml{g}) \}$. 
	For every $\mlc \in \N_\Gl$, we define the knowledge base associated to the context as

\medskip
\centerline{\small
    $\KB_\mlc = \bigcup\{\KB_\mlm \in \CKB \;|\; PG(\Gl) \models \triple(\mlc,\mlmod,\mlm,\ml{g}) \}$}
\medskip
\item
  We define each \emph{local program} $PC(\mlc)$ for $\mlc \in \N_\Gl$ by 
	applying input rules $I_{loc}$ and $I_{rl}$ to $\KB_\mlc$
	and adding deduction rules $P_{loc}$ and $P_{rl}$.
\item
  The final \emph{CKR program} $PK(\CKB)$ is then defined as
	the union of $PG(\Gl)$ with all local programs $PC(\mlc)$.
\end{enumerate}
We say that $\CKB$ \emph{entails} an axiom $\alpha$ 
in a context $\mlc \in \N$ %(denoted $\CKB \dimp \mlc:\alpha$),
if the elements of $PK(\CKB)$ and $O(\alpha, \ml{c})$ are defined and
$PK(\CKB) \models O(\alpha, \ml{c})$.
We can show (see~\cite{BozzatoSerafini:13}) that the presented rules and 
translation process provide a sound and complete calculus 
for instance checking for $\SROIQrl$ CKR.

%###################################

%-----------------------------------------------------------------------------  

%  \input{sec/Springles}

%-----------------------------------------------------------------------------
\section{CKR Implementation on RDF}
\label{sec:springles}

% # CKR prototype: intro #
We recently presented a prototype~\cite{BozzatoSerafini:13}
implementing the forward reasoning procedure over CKR expressed by the 
materialization calculus.
The prototype accepts RDF input data expressing OWL-RL axioms and assertions
for global and local knowledge modules: these different pieces of knowledge
are represented as distinct named graphs, while contextual primitives 
have been encoded in a RDF vocabulary.
The prototype is based on an extension of the Sesame RDF Framework\footnote{\url{http://www.openrdf.org/}}
and structured in a client-server architecture: 
the main component, called \emph{CKR core} module and residing in the server-side part,
exposes the CKR primitives and a SPARQL 1.1 endpoint for query and update operations on the contextualized knowledge.
The module offers the ability to compute and materialize the inference closure of the input CKR,
add and remove knowledge and execute queries over the complete CKR structure.
%\lbnote{Figure?}

% # SPRINGLES (sketch) #
%[What is particular for this implementation (the CKR core module) 
%is the ability to compute inferences
%over multiple named graphs.]
The distribution of knowledge in different named graphs asks for a 
component to compute inference over multiple graphs in a RDF store,
since inference mechanisms in current stores usually ignore the graph part.
This component has been realized as a general software layer
called 
\emph{SPRINGLES}\footnote{\emph{SParql-based Rule Inference over Named Graphs Layer Extending Sesame}.}.
Intuitively, the layer provides methods to demand a 
closure materialization on the RDF store data: rules are encoded as named graphs aware
SPARQL queries and it is possible to customize both the input ruleset and the evaluation strategy.
The general form of SPRINGLES rules is the following:
\begin{lstlisting}[breaklines=true,language=ttl]
:<rule-name> a spr:Rule ;
  spr:head """ <graph pattern>""" ;
  spr:body """ <sparql query>""" .
\end{lstlisting}
\noindent
\verb|<graph-pattern>| is an RDF (named) graph that can contain a set
of variables, which are bounded in the SPARQL query of the body. The
body of a rule is a SPARQL query that is evaluated. The result of the
evaluation of the rule body is a set of bindings for the variables
that occurs in the rule head. For every such a binding the
corresponding statement in the head of the rule is added to the
repository. 

% # CKR calculus in SPRINGLES #
In our case, the ruleset basically encodes the rules of the 
presented materialization calculus. % in SPARQL-based forward rules.
% # Esempio di regola SUBC #
As an example, we present the rule dealing with atomic concept inclusions:

%[REGOLA PER SUBC]
\begin{lstlisting}[breaklines=true,language=ttl]
:prl-subc a spr:Rule ;
  spr:head """ GRAPH ?mx { ?x rdf:type ?z } """ ;
  spr:body """ GRAPH ?m1 { ?y rdfs:subClassOf ?z }
               GRAPH ?m2 { ?x rdf:type ?y }
               GRAPH sys:dep { ?mx sys:derivedFrom ?m1,?m2 }
               FILTER NOT EXISTS 
                 { GRAPH ?m0 { ?x rdf:type ?z }
                   GRAPH sys:dep { ?mx sys:derivedFrom ?m0 } } """ .
\end{lstlisting}
where prefix \verb|spr:| corresponds to symbols in the vocabulary of
SPRINGLES objects and \verb|sys:| prefixes utility ``system'' symbols used in the 
definition of the rules evaluation plan.
Intuitively, when the condition in the body part of the rule is
verified in graphs {\tt ?m1} and {\tt ?m2}, the head part is
materialized in the inference graph {\tt ?mx}.  Note that in the
formulation of the rule we work at the level of knowledge modules
(i.e. named graphs). Note that the body of the
rules contains a ``filter'' condition, which is a SPARQL based method
to avoid the duplication of conclusions: the \verb|FILTER| 
condition imposes a rule to be fired only if its conclusion is not
already present in the context.

%This code corresponds to the local rule (prl-subc) of $P_{rl}$, thus has the scope of a single context.
%When the condition in the body part of the rule is verified
%in graphs {\tt ?m1} and {\tt ?m2}, the head part is materialized in the inference graph {\tt ?mx}.
%In the rule we work at level of knowledge modules (i.e. named graphs): 
%the first three lines directly correspond to the rule in $P_{rl}$;
%in the fourth line we require that module {\tt ?mx}, containing the inferences in 
%the given context, 
%%and created in the initial closure of the global context,
%depends on the modules of the assumptions. 
%In other words, we require that both rule preconditions and 
%results belong to the KB associated to 
%the same context\footnote{Statements {\tt ?mx sys:derivedFrom ?my} are generated
%%by an auxiliary rule 
%from the closure of the global context. For every context,
%an additional module for its inferences is created and associated to it via 
%{\tt sys:derivedFrom} relation.}.
%The filter expression checks that the results have not been derived yet.

% # Evaluation strategy #
%As previously mentioned, such
The rules are evaluated with a %respect to a 
strategy that basically follows the same steps of the
translation process defined for the calculus.
%SPRINGLES plan specification language allows to define evaluation
%strategies by means of primitives for parallel execution, sequence, fixpoint and iteration.
%Intuitively, %(the actual plan can be found in the prototype demo), 
The plan goes as follows:
(i) we compute the closure on the graph for %representing the
global context $\Gl$, by a fixpoint on rules corresponding to $P_{rl}$;
(ii) we derive associations between contexts and their modules,
by adding dependencies %in the graph {\tt sys:dep} 
for every assertion of the kind $\mlmod(\mlc, \mlm)$ 
in the global closure;
(iii) we compute the closure the contexts, by applying rules encoded from $P_{rl}$ and $P_{loc}$ and
resolving $eval$ expressions by the metaknowledge information in the global closure.

% # Demo: link a prototipo #
%A demo of the prototype, containing RDF data that encodes the example CKR $\CKBt$,
%can be found at \url{https://dkm.fbk.eu/index.php/CKR-TourismDemo}.

\vspace{-1ex}

%  \input{sec/experiments}

%-----------------------------------------------------------------------------
\section{Experimental Evaluation}
\label{sec:experiments}

In this section we illustrate the experiments we performed to assess the performance
of the CKR prototype and their results.
We begin by presenting the method we used to create the synthetic test sets 
that we generated for such evaluation.

%-  Description of the generation of synthetic datasets
%- - - - - - - - - - - - - - - - - - - - - - - - - - - - - - - - - - - - - - -
\smallskip\noindent
\textbf{Generation of synthetic test sets.}
%
%We performed an evaluation of the MetaReasons prototype with respect to 
%scalability of reasoning in OWL-RL:
%the test sets of UKBs over which such evaluation is performed 
%have been synthetically generated.
In order to create our test sets, we developed a simple generator
that can output randomly generated CKRs with certain features.
In particular, for each generated CKR, the generator takes in input: 
%\begin{itemize}
%\item 
(1) the number $n$ of contexts (i.e. local named graphs) to be generated;
%\item   
(2) the dimensions of the signature to be declared 
	   (number $m$ of base classes, $l$ of properties and $k$ of individuals);
%\item
(3) the axiom size for the global and local modules 
   (number of global TBox, ABox and RBox axioms and
   number of TBox, ABox and RBox axioms per context);
%\item
(4) optionally, the number of additional local $\eval$ axioms and 
    the number of individuals to be propagated across contexts.	
%\end{itemize}
%The generated CKRs are then saved as TRIG files, 
%that can be readily imported in the RDF store for MetaReasons.
%
Intuitively, the generation of a CKR proceeds as follows:
\begin{enumerate}
\item 
  The contexts (named ${\tt :\!c0}, \dots, {\tt :\!cn}$) are declared in the global context named graph
	and are linked to a different module name (${\tt :\!m0}, \dots, {\tt :\!mn}$), 
	corresponding to the named graph containing their local knowledge.
\item
  Base classes (named ${\tt :\!A0}, \dots, {\tt :\!Am}$), 
  object properties (${\tt :\!R0}, \dots, {\tt :\!Rl}$) and individuals 
  (${\tt :\!a0}, \dots, {\tt :\!ak}$) 
  are added to the global graph: these symbols are used in the generation of 
	global and local axioms.
\item
  Then generation of global axioms takes place. We chose to generate axioms as follows, 
	in order to create realistic instances of knowledge bases:
  \begin{itemize}
	\item 
    Classes and properties names are taken from the base signature using random selection criteria
    in the form of (the positive part of) a Gaussian curve centered in $0$: 
    intuitively, classes equal or near to ${\tt :\!A0}$ are more probable %to appear 
		in axioms than ${\tt :\!An}$.
	\item 
	  Individuals are randomly selected using a uniform distribution.
	\item 
	  TBox, ABox and RBox axioms in $\SROIQrl$ are added in the requested number 
		to the global context module following the percentages shown in Table~\ref{tab:percentage} 
	  (note that the reported axioms are normal form $\SROIQrl$ axioms, as defined in~\cite{BozzatoSerafini:13}).
	  Such percentages have been selected in order to simulate the common distribution in the use of 
	  the constructs in real knowledge bases.
  \end{itemize}
\item
  The same generation criteria are then applied in the case of local graphs representing the 
	local knowledge of contexts.
\item
  If specified, the requested number for $\eval$ axioms of the form $\eval(A, \ml{C}) \subs B$
	and for the set of individuals in the scope of the $\eval$ operator (i.e. as local members of $A$)
	are added to local contexts graphs.
\end{enumerate}

\begin{table}[t]
\vspace{-2ex}
\begin{center}\scalebox{0.82}{
\begin{tabular}{|r|c|}
\hline
  {\bf TBox axiom} & {\bf \%} \\
\hline  
   $A \subs B$ & 50\% \\
   $A \subs \non B$ & 20\% \\   
   $A \subs \exists R.\{a\}$ & 10\% \\
   $A \sqcap B \subs C$ & 5\% \\   
   $\exists R.A \subs B$ & 5\% \\      
   $A \subs \forall R.B$ & 5\% \\   
   $A \subs \leqslant\!1 R.B$ & 5\% \\      
\hline      
\end{tabular}
\qquad
\begin{tabular}{|r|c|}
\hline
  {\bf ABox axiom} & {\bf \%} \\
\hline  
   $A(a)$ & 40\% \\
   $R(a,b)$ & 40\% \\   
   $\non R(a,b)$ & 10\% \\      
   $a = b$ & 5\% \\      
   $a \neq b$ & 5\% \\         
\hline      
\end{tabular}
\qquad
\begin{tabular}{|r|c|}
\hline
  {\bf RBox axiom} & {\bf \%} \\
\hline  
   $R \subs T$ & 50\% \\
   $\mathrm{Inv}(R,S)$ & 25\% \\   
   $R \circ S \subs T$ & 10\% \\   
   $\mathrm{Dis}(R,S)$ & 10\% \\   
   $\mathrm{Irr}(R)$ & 5\% \\
   \hline      
\end{tabular}}
\end{center}
  \caption{Percentages of generated axioms.}%
  \label{tab:percentage}
\vspace{-5.5ex}	
\end{table}

%\lbnote{CONTINUARE: separare in CKR1 e CKR2 e spostare la seguente descrizione e tabella in CKR1 
%+ spostare qui la evaluation setup!}

%-  Experimental setup 
%- - - - - - - - - - - - - - - - - - - - - - - - - - - - - - - - - - - - - - -
\smallskip\noindent
\textbf{Experimental Setup.}
Evaluation experiments were carried out on a 4 core Dual Intel Xeon Processor
machine with 32Gb 1866MHZ DDR3 RAM, standard S-ATA (7.200RPM) HDD, running 
a Linux RedHat 6.5 distribution. 
We allocated 6Gb of memory to the JVM running the SPRINGLES web-app
(i.e. the RDF storage and inference prototype), while
20Gb were allocated to the utility program managing the upload, profiling and 
cleaning of the test repositories.
In order to abstract from the possible overhead 
for the repository setup, the tests have been averaged over %5 runs 
multiple runs of the closure operation for each CKR. 
%Moreover, the tests results have been averaged over 3 versions of test sets
%in order to reduce the impact of special cases in the random generation.

The tests were carried out on different CKR rulesets in order to 
study their applicability in practical reasoning.
The rulesets are limitations to the full set of rules
and evaluation strategy presented in previous sections, in particular:
\begin{itemize}
\item 
  \emph{ckr-rdfs-global:}
  inference is only applied to the global context (no local reasoning inside 
  local contexts named graphs). Applies only inference rules 
  for RDFS and for the definition of CKR structure (e.g. association of 
	named graphs for knowledge modules to contexts).
\item 
  \emph{ckr-rdfs-local:}  
  inference is applied to the graphs both for global and local contexts. 
  Again, applies only inference rules for RDFS and CKR structure rules.  
\item 
  \emph{ckr-owl-global:}  
  inference is only applied to the global context, considering all of the 
  inference rules for $\SROIQrl$ and CKR structure rules.
\item 
  \emph{ckr-owl-local:}    
  full strategy defined by the materialization calculus.
  Inference is applied to the global and local parts,
  using all of the (global and local) $\SROIQrl$ and CKR rules.
\end{itemize}
More in details, application of RDFS rules corresponds to the limitation of 
  OWL RL closure step only to the inference rules for 
	subsumption on classes %(rule \emph{prln-subc}) 
	and object properties. 
  %(rule \emph{prln-subr}).
%
%In order to assess the impact of the RDF storage solution to the system perfomances,
%we performed the experiments using as back-ends both \emph{OWLIM lite}\footnote{\url{http://www.ontotext.com/owlim}} 
%and \emph{Sesame Native}\footnote{\url{http://www.openrdf.org/}} RDF stores.

%- - - - - - - - - - - - - - - - - - - - - - - - - - - - - - - - - - - - - - -
\smallskip\noindent
\textbf{TS1: scalability evaluation.}
%
%We carried out experiments on previously presented test set
The first experiments we carried out on the CKR prototype
had the task to determine the (average) inference closure time %and repository dimensions
with respect to the increase in number of contexts and their contents:
with reference to the research questions in the introduction, 
this first evaluation aimed at answering question \textbf{RQ1}.

Using the CKR generator tool, we generated the set of test CKRs
%The generated test sets are 
shown in Table~\ref{tab:TS1}: we call this test set \emph{TS1}.
\begin{table}[tp]
\vspace{-2ex}
\scriptsize\centering
\begin{tabular}{|r|rrr|rrr|rrr|r|}
\hline
& &&& \multicolumn{2}{l}{{\bf Global KB}} && \multicolumn{2}{l}{{\bf Local KBs}} && \\
\hline
{\bf Contexts}	&	{\bf Classes}	&	{\bf Roles}	&	{\bf Indiv.}	&	
{\bf TBox}	&	{\bf RBox}	&	{\bf ABox}	&	
{\bf TBox}	&	{\bf RBox}	&	{\bf ABox}	&	{\bf Total axioms}	\\
\hline
1	&	10	&	10	&	20	&	10	&	5	&	20	&	10	&	5	&	20	&	70	\\
1	&	50	&	50	&	100	&	50	&	25	&	100	&	50	&	25	&	100	&	350	\\
1	&	100	&	100	&	200	&	100	&	50	&	200	&	100	&	50	&	200	&	700	\\
1	&	500	&	500	&	1000	&	500	&	250	&	1000	&	500	&	250	&	1000	&	3.500	\\
1	&	1000	&	1000	&	2000	&	1000	&	500	&	2000	&	1000	&	500	&	2000	&	7.000	\\
%1	&	5000	&	5000	&	10000	&	5000	&	2500	&	10000	&	5000	&	2500	&	10000	&	35.000	\\
%1	&	10000	&	10000	&	20000	&	10000	&	5000	&	20000	&	10000	&	5000	&	20000	&	70.000	\\
%1	&	50000	&	50000	&	100000	&	50000	&	25000	&	100000	&	50000	&	25000	&	100000	&	350.000	\\
\hline
5	&	10	&	10	&	20	&	10	&	5	&	20	&	10	&	5	&	20	&	210	\\
5	&	50	&	50	&	100	&	50	&	25	&	100	&	50	&	25	&	100	&	1.050	\\
5	&	100	&	100	&	200	&	100	&	50	&	200	&	100	&	50	&	200	&	2.100	\\
5	&	500	&	500	&	1000	&	500	&	250	&	1000	&	500	&	250	&	1000	&	10.500	\\
5	&	1000	&	1000	&	2000	&	1000	&	500	&	2000	&	1000	&	500	&	2000	&	21.000	\\
%5	&	5000	&	5000	&	10000	&	5000	&	2500	&	10000	&	5000	&	2500	&	10000	&	105.000	\\
%5	&	10000	&	10000	&	20000	&	10000	&	5000	&	20000	&	10000	&	5000	&	20000	&	210.000	\\
%5	&	50000	&	50000	&	100000	&	50000	&	25000	&	100000	&	50000	&	25000	&	100000	&	1.050.000	\\
\hline
10	&	10	&	10	&	20	&	10	&	5	&	20	&	10	&	5	&	20	&	385	\\
10	&	50	&	50	&	100	&	50	&	25	&	100	&	50	&	25	&	100	&	1.925	\\
10	&	100	&	100	&	200	&	100	&	50	&	200	&	100	&	50	&	200	&	3.850	\\
10	&	500	&	500	&	1000	&	500	&	250	&	1000	&	500	&	250	&	1000	&	19.250	\\
10	&	1000	&	1000	&	2000	&	1000	&	500	&	2000	&	1000	&	500	&	2000	&	38.500	\\
%10	&	5000	&	5000	&	10000	&	5000	&	2500	&	10000	&	5000	&	2500	&	10000	&	192.500	\\
%10	&	10000	&	10000	&	20000	&	10000	&	5000	&	20000	&	10000	&	5000	&	20000	&	385.000	\\
%10	&	50000	&	50000	&	100000	&	50000	&	25000	&	100000	&	50000	&	25000	&	100000	&	1.925.000	\\
\hline
50	&	10	&	10	&	20	&	10	&	5	&	20	&	10	&	5	&	20	&	1.785	\\
50	&	50	&	50	&	100	&	50	&	25	&	100	&	50	&	25	&	100	&	8.925	\\
50	&	100	&	100	&	200	&	100	&	50	&	200	&	100	&	50	&	200	&	17.850	\\
50	&	500	&	500	&	1000	&	500	&	250	&	1000	&	500	&	250	&	1000	&	89.250	\\
50	&	1000	&	1000	&	2000	&	1000	&	500	&	2000	&	1000	&	500	&	2000	&	178.500	\\
%50	&	5000	&	5000	&	10000	&	5000	&	2500	&	10000	&	5000	&	2500	&	10000	&	892.500	\\
%50	&	10000	&	10000	&	20000	&	10000	&	5000	&	20000	&	10000	&	5000	&	20000	&	1.785.000	\\
%50	&	50000	&	50000	&	100000	&	50000	&	25000	&	100000	&	50000	&	25000	&	100000	&	8.925.000	\\
\hline
100	&	10	&	10	&	20	&	10	&	5	&	20	&	10	&	5	&	20	&	3.535	\\
100	&	50	&	50	&	100	&	50	&	25	&	100	&	50	&	25	&	100	&	17.675	\\
100	&	100	&	100	&	200	&	100	&	50	&	200	&	100	&	50	&	200	&	35.350	\\
100	&	500	&	500	&	1000	&	500	&	250	&	1000	&	500	&	250	&	1000	&	176.750	\\
100	&	1000	&	1000	&	2000	&	1000	&	500	&	2000	&	1000	&	500	&	2000	&	353.500	\\
%100	&	5000	&	5000	&	10000	&	5000	&	2500	&	10000	&	5000	&	2500	&	10000	&	1.767.500	\\
%100	&	10000	&	10000	&	20000	&	10000	&	5000	&	20000	&	10000	&	5000	&	20000	&	3.535.000	\\
%100	&	50000	&	50000	&	100000	&	50000	&	25000	&	100000	&	50000	&	25000	&	100000	&	17.675.000	\\
%\hline
%500	&	10	&	10	&	20	&	10	&	5	&	20	&	10	&	5	&	20	&	17.535	\\
%500	&	50	&	50	&	100	&	50	&	25	&	100	&	50	&	25	&	100	&	87.675	\\
%500	&	100	&	100	&	200	&	100	&	50	&	200	&	100	&	50	&	200	&	175.350	\\
%500	&	500	&	500	&	1000	&	500	&	250	&	1000	&	500	&	250	&	1000	&	876.750	\\
%500	&	1000	&	1000	&	2000	&	1000	&	500	&	2000	&	1000	&	500	&	2000	&	1.753.500	\\
%500	&	5000	&	5000	&	10000	&	5000	&	2500	&	10000	&	5000	&	2500	&	10000	&	8.767.500	\\
%500	&	10000	&	10000	&	20000	&	10000	&	5000	&	20000	&	10000	&	5000	&	20000	&	17.535.000	\\
%500	&	50000	&	50000	&	100000	&	50000	&	25000	&	100000	&	50000	&	25000	&	100000	&	87.675.000	\\
%\hline
%1000	&	10	&	10	&	20	&	10	&	5	&	20	&	10	&	5	&	20	&	35.035	\\
%1000	&	50	&	50	&	100	&	50	&	25	&	100	&	50	&	25	&	100	&	175.175	\\
%1000	&	100	&	100	&	200	&	100	&	50	&	200	&	100	&	50	&	200	&	350.350	\\
%1000	&	500	&	500	&	1000	&	500	&	250	&	1000	&	500	&	250	&	1000	&	1.751.750	\\
%1000	&	1000	&	1000	&	2000	&	1000	&	500	&	2000	&	1000	&	500	&	2000	&	3.503.500	\\
%1000	&	5000	&	5000	&	10000	&	5000	&	2500	&	10000	&	5000	&	2500	&	10000	&	17.517.500	\\
%1000	&	10000	&	10000	&	20000	&	10000	&	5000	&	20000	&	10000	&	5000	&	20000	&	35.035.000	\\
%1000	&	50000	&	50000	&	100000	&	50000	&	25000	&	100000	&	50000	&	25000	&	100000	&	175.175.000	\\
\hline
\end{tabular}
\vspace{2ex}
\caption{Test set TS1.}
\label{tab:TS1}
\vspace{-8ex}
\end{table}
Intuitively, TS1 
%\begin{itemize}
%\item 
  contains sets of CKRs with an increasing number of contexts,
  in which CKRs have an increasing number of axioms.
  We note that no $\eval$ axioms were added to TS1 knowledge bases.
  %Aim of this set is to simply assess how the system performs in presence of an 
  %increase both in number and in dimension of local knowledge bases.	
	%number of datasets and axioms.
%\item
  %in test set MR2, we generated again sets of UKBs with an increasing number of dataset,
  %but we fixed the total number of local axioms.
  %Goal of this set is to show how the system performs with different distribution of knowledge across the datasets.
%\end{itemize}
%In order to smooth the influence of anomalies given by particular
%cases in the randomly generated axioms, we performed our tests over 
%4 instances of generation of testset MR1.

%\subsection{Evaluation results}
%-  Results of the evaluation (for different rulesets and storage solutions)  
%-  (Comments on the results)
We ran the CKR prototype on 3 generations of TS1
also varying the reasoning regime among the rulesets detailed above:
the different generation instances of TS1 are necessary
in order to reduce the impact of special cases in the random generation.
The results of the experiments
on TS1 are reported in Table~\ref{tab:TS1results}.
In the table, for each of the generated CKRs (referred by number of contexts and 
number of base classes in the first two columns), we show the number of
total asserted triples in column \emph{Triples} (averaged on the 3 versions of TS1).
The following columns list the results
%\footnote{
  %Values as the ones grayed out in the tables are due
	%to particular cases in the generated axioms: 
	%these must be considered as anomalies and have been left out from the graphs.}
of the closure for each of the rulesets: for a ruleset, we list the (average) total number
of triples (asserted + inferred), the inferred triples and the (average) time in milliseconds
for the closure operation.
The value \emph{timedout} in the measures indicates that the closure operation
exceeded 30 minutes (1.800.000 ms.).

\begin{table}[tp]
\vspace{-2ex}
\scriptsize\hspace{-9ex}
\begin{tabular}{|r@{\;\;\;}r|r|r@{\;\;\;}r@{\;\;\;}r|r@{\;\;\;}r@{\;\;\;}r|r@{\;\;\;}r@{\;\;\;}r|r@{\;\;\;}r@{\;\;\;}r|}
\hline
&&& \multicolumn{2}{l}{{\bf ckr-rdfs-global}} 
 && \multicolumn{2}{l}{{\bf ckr-owl-global}} 
 && \multicolumn{2}{l}{{\bf ckr-rdfs-local}} 
 && \multicolumn{2}{l}{{\bf ckr-owl-local}} &\\
\hline
{\bf Ctx.}	&	{\bf Cls.}	&	{\bf Triples}	&
{\bf Total}	&	{\bf Inf.}	&	{\bf Time}	&	
{\bf Total}	&	{\bf Inf.}	&	{\bf Time}	&	
{\bf Total}	&	{\bf Inf.}	&	{\bf Time}	&	
{\bf Total}	&	{\bf Inf.}	&	{\bf Time}	\\
\hline
1	&	10	&	208	&	228	&	20	&	222	&	234	&	26	&	326	&	249	&	41	&	291	&	298	&	90	&	868	\\
1	&	50	&	1079	&	1165	&	87	&	221	&	1288	&	209	&	518	&	1351	&	272	&	323	&	1918	&	839	&	4596	\\
1	&	100	&	2165	&	2398	&	233	&	260	&	2666	&	501	&	943	&	2687	&	521	&	346	&	3803	&	1638	&	15916	\\
1	&	500	&	10549	&	11870	&	1321	&	846	&	13293	&	2743	&	22930	&	14833	&	4284	&	2461	&	22828	&	12278	&	556272	\\
1	&	1000	&	20981	&	23600	&	2619	&	1528	&	25957	&	4976	&	95957	&	29993	&	9012	&	4644	&	\gray{timedout} &	\gray{timedout}	&	\gray{timedout} \\
\hline
5	&	10	&	644	&	685	&	41	&	176	&	698	&	54	&	226	&	780	&	136	&	193	&	1470	&	826	&	11721	\\
5	&	50	&	3124	&	3259	&	135	&	190	&	3330	&	205	&	341	&	4134	&	1010	&	522	&	9874	&	6750	&	328107	\\
5	&	100	&	6201	&	6450	&	249	&	254	&	6675	&	475	&	962	&	8845	&	2645	&	1258	&	31615	&	25414	&	913617	\\
5	&	500	&	30928	&	31994	&	1066	&	719	&	33025	&	2097	&	23109	&	44987	&	14059	&	7819	&	\gray{timedout}	&	\gray{timedout}	&	\gray{timedout}	\\
5	&	1000	&	61691	&	64363	&	2672	&	1491	&	66661	&	4969	&	106967	&	95636	&	33945	&	16291	&	\gray{timedout}	&	\gray{timedout}	&	\gray{timedout}\\
\hline
10	&	10	&	1149	&	1216	&	66	&	165	&	1225	&	76	&	202	&	1427	&	278	&	541	&	6141 & 4992	&	448249 \\
10	&	50	&	5620	&	5782	&	163	&	210	&	5895	&	275	&	460	&	8008	&	2388	&	1392	&	\gray{timedout}	&	\gray{timedout}	&	\gray{timedout}	\\
10	&	100	&	11058	&	11353	&	295	&	281	&	11865	&	807	&	1745	&	16315	&	5257	&	2986	&	\gray{timedout}	&	\gray{timedout}	&	\gray{timedout}	\\
10	&	500	&	56578	&	57836	&	1258	&	910	&	59052	&	2474	&	33643	&	86821	&	30243	&	17375	&	\gray{timedout}	&	\gray{timedout}	&	\gray{timedout}	\\
10	&	1000	&	112824	&	115273	&	2449	&	2030	&	117666	&	4842	&	114443	&	173938	&	61113	&	36647	&	\gray{timedout}	&	\gray{timedout}	&	\gray{timedout}	\\
\hline
50	&	10	&	5509	&	5780	&	271	&	208	&	5785	&	276	&	256	&	7003	&	1494	&	2167	&	\gray{timedout}	&	\gray{timedout}	&	\gray{timedout}	\\
50	&	50	&	26327	&	26676	&	348	&	323	&	26795	&	467	&	825	&	35640	&	9312	&	14598	&	\gray{timedout}	&	\gray{timedout}	&	\gray{timedout}	\\
50	&	100	&	52037	&	52543	&	506	&	603	&	52749	&	713	&	2384	&	78439	&	26402	&	21461	&	\gray{timedout}	&	\gray{timedout}	&	\gray{timedout}	\\
50	&	500	&	259810	&	261355	&	1546	&	2025	&	262722	&	2913	&	41973	&	416088	&	156278	&	299504	&	\gray{timedout}	&	\gray{timedout}	&	\gray{timedout}	\\
50	&	1000	&	520276	&	523082	&	2807	&	4350	&	525702	&	5426	&	214434	&	827451	&	307176	&	397110	&	
\gray{timedout}	&	\gray{timedout}	&	\gray{timedout}	\\
\hline
100	&	10	&	10658	&	11171	&	513	&	242	&	11181	&	523	&	279	&	12916	&	2258	&	1865	&	\gray{timedout}	&	\gray{timedout}	&	\gray{timedout}	\\
100	&	50	&	51709	&	52347	&	638	&	442	&	52461	&	752	&	1241	&	73639	&	21930	&	31003	&	\gray{timedout}	&	\gray{timedout}	&	\gray{timedout}	\\
100	&	100	&	103341	&	104035	&	694	&	531	&	104259	&	918	&	2784	&	145788	&	42447	&	47179	&	\gray{timedout}	&	\gray{timedout}	&	\gray{timedout}	\\
100	&	500	&	514497	&	516316	&	1819	&	3469	&	517567	&	3070	&	87325	&	844215	&	329718	&	774657	&	\gray{timedout}	&	\gray{timedout}	&	\gray{timedout}	\\
100	&	1000	&	1028233	&	1031367	&	3135	&	7835	&	1033725	&	5492	&	394881	&	1674765	&	646532	&	1018616	&	\gray{timedout}	&	\gray{timedout}	&	\gray{timedout}	\\
\hline
\end{tabular}
\vspace{0.7ex}
\caption{Scalability results for test set TS1.}
\label{tab:TS1results}
\vspace{-5ex}
\end{table}

\begin{figure}[tp]
%\vspace{-4ex}
\hspace{-15ex}
%\centering
%\rotatebox{0}{
\begin{tabular}{c@{\;\;\;\;}c}
\includegraphics[width=.65\columnwidth]{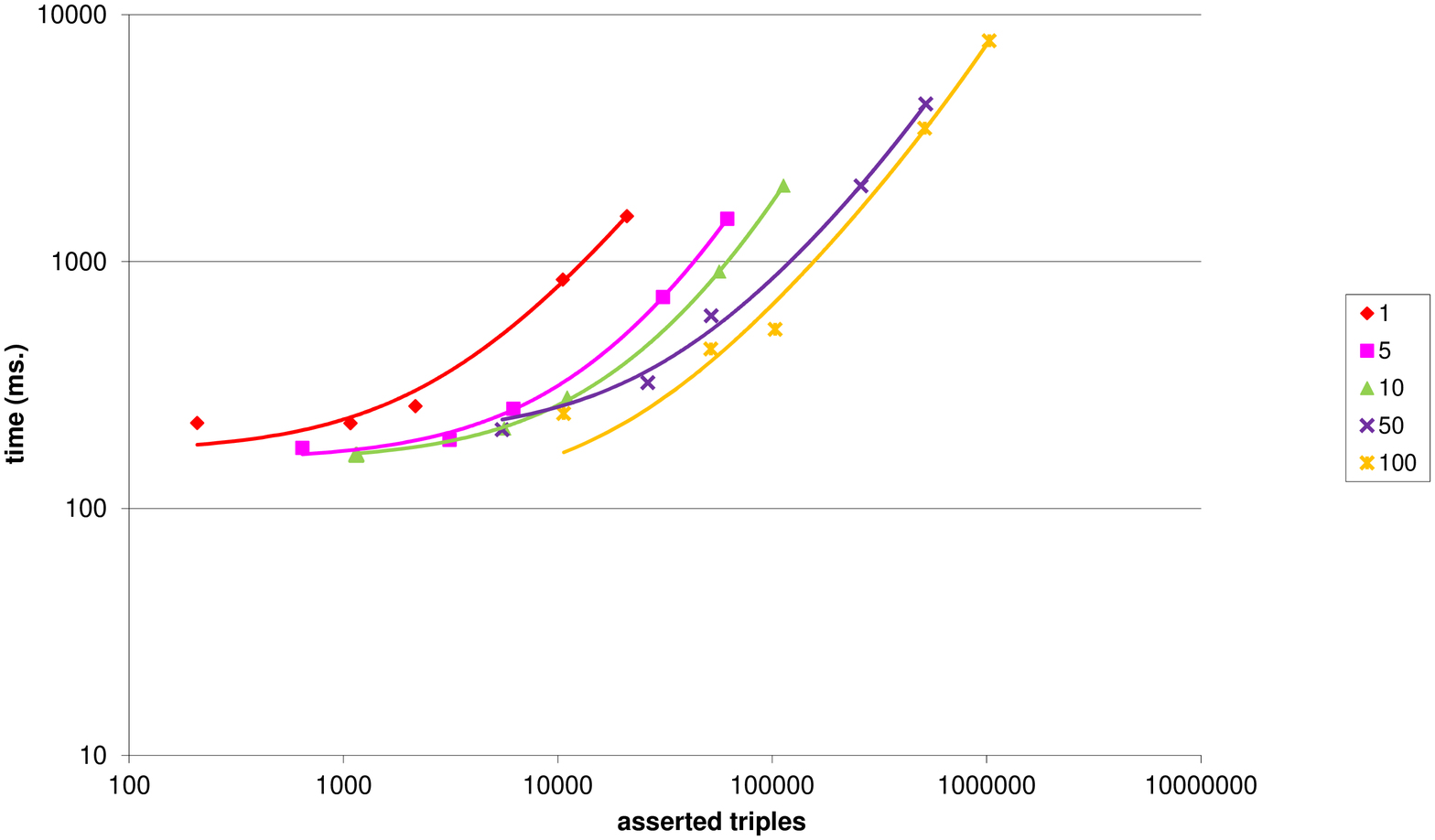} &
\includegraphics[width=.65\columnwidth]{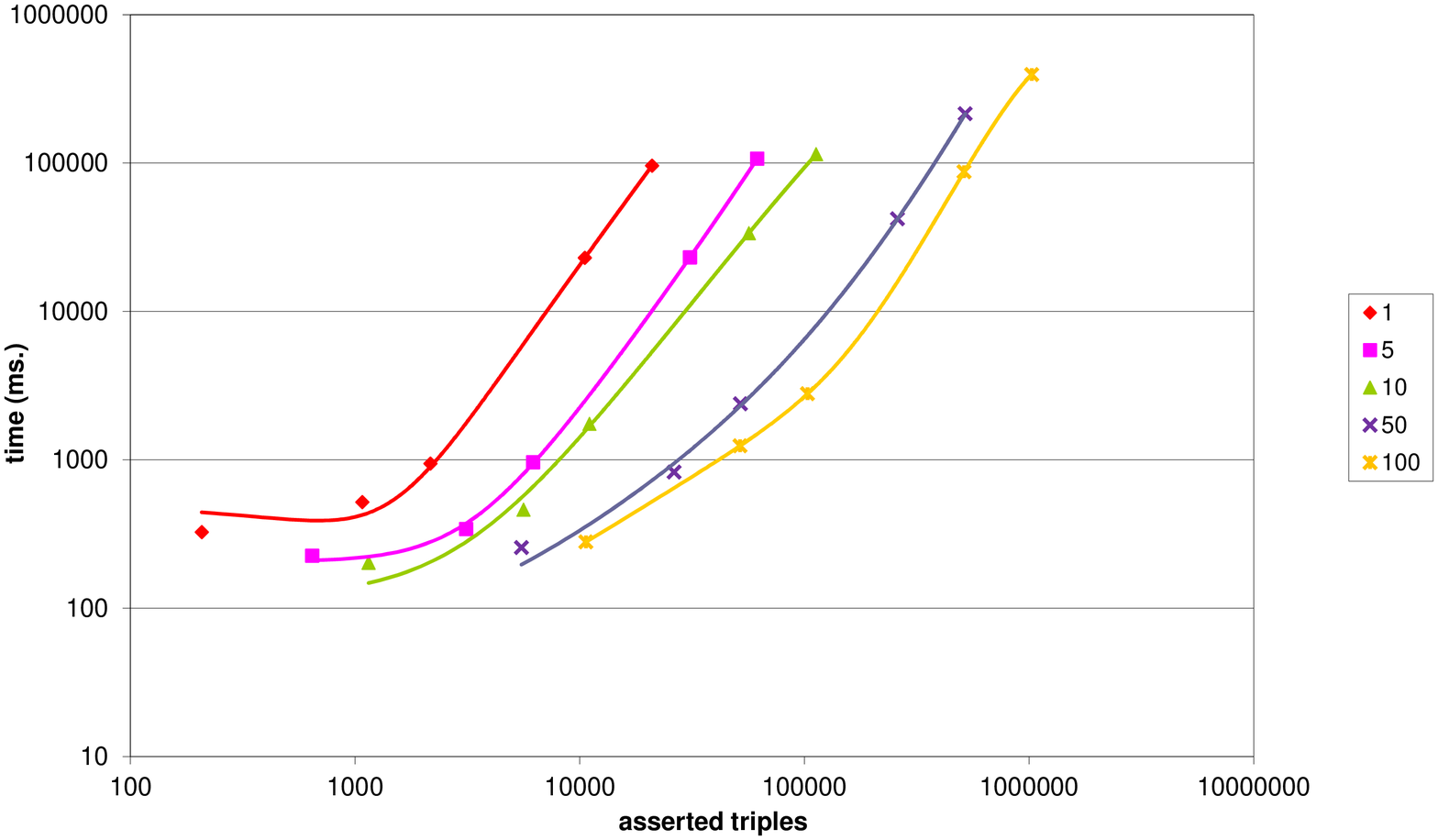}\\
a) ckr-rdfs-global &
b) ckr-owl-global\\[2ex]
\includegraphics[width=.65\columnwidth]{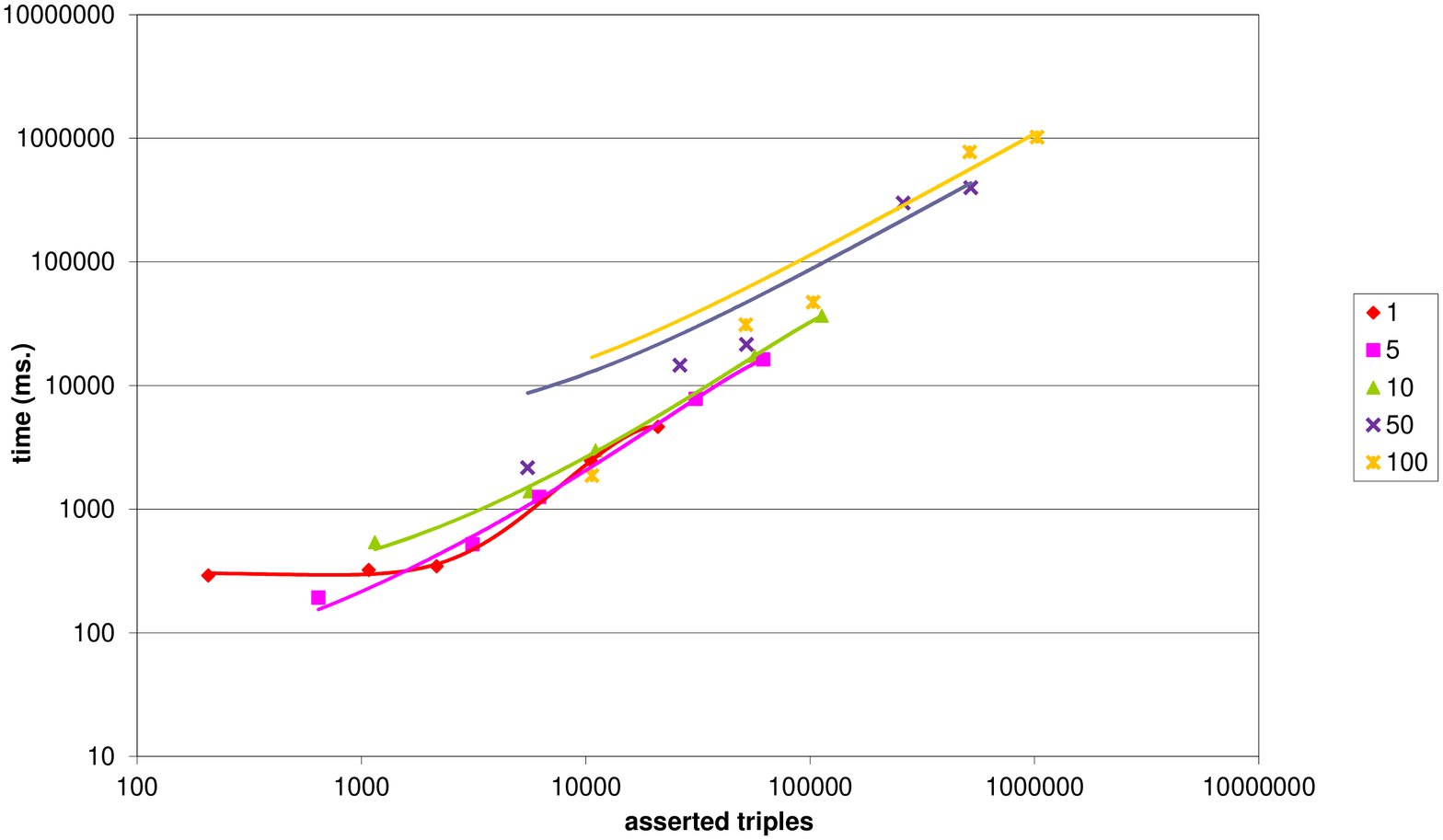} &
\includegraphics[width=.65\columnwidth]{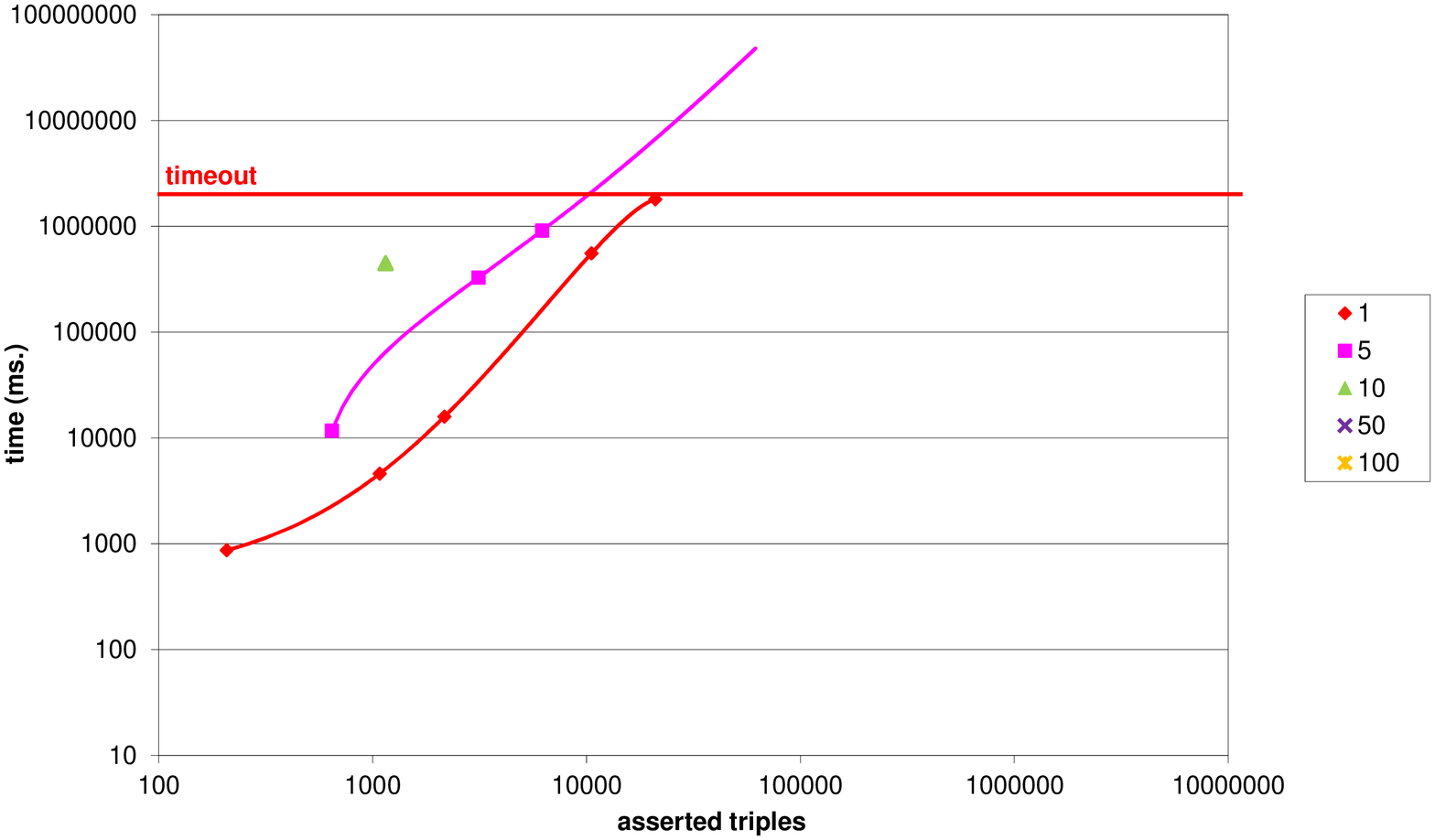}\\
c) ckr-rdfs-local &
d) ckr-owl-local
\end{tabular}
%}
\vspace{-3ex}
\caption{Scalability graphs for TS1.}%
\label{fig:graph-Sesame}%
\end{figure}

In order to analyze the results, the behaviour of the prototype for each
of the rulesets has been plotted to graphs, shown in Figure~\ref{fig:graph-Sesame}.
%\lbreview{AGGIORNARE DATI e GRAFI PER TS1!}
%in Figure~\ref{fig:graph-OWLIM} 
%it is shown the case for the OWLIM back-end and 
%in Figure~\ref{fig:graph-Sesame} for Sesame Native storage.
Each of the series represents a set with a fixed number of contexts (1 to 100)
and each point a CKR. The $x$ axis represents the number of asserted triples, while the
$y$ axis shows the time in milliseconds; the red horizontal line depicts the 30 minutes limit for timeout.
To better visualize the behaviour of the series, we plotted a trend line
for each of the series: the lines represent an approximation of the data trend
calculated by polynomial regression\footnote{Average $R^2$ value across all approximations
is $\geq 0,993$.}.
%\lbreview{AGGIORNARE $R^2$ PER TS1}

Some conclusions can be derived from these data and graphs: 
the first most evident fact is that the reasoning regime strongly impacts
the scalability of the system.
Thus, in practical cases the choice of a naive application
of the full OWL RL ruleset might not be viable, in presence of 
large local datasets: on the other hand, 
if expressive reasoning inside contexts is not required,
scalability can be enhanced by relying on the RDFS rulesets
(or, in general, by carefully tailoring the ruleset to the required expressivity).

%On the other hand, the management of the named graphs separation
%of knowledge (with equal number of triples) does not seem to directly 
%impact on the scalability of the prototype.
By analyzing the graphs and the approximations, 
it is also possible to observe that the system shows a different behaviour 
depending on the different reasoning regimes.
In the case of \emph{ckr-rdfs-global} and \emph{ckr-owl-global},
the results suggest that the management of named graphs 
does not add overhead to the reasoning in the global context.
This can be also seen by checking Table~\ref{tab:TS1results}: %it is possible to see that
for a similar number of inferred triples
the separation across different graphs does not influence the 
reasoning time. For example, this is visible 
for cases with similar $y$ values of the graph (e.g. the case for 1000 classes
in series for 1 and 5 contexts, in both rulesets).
In the case of \emph{ckr-rdfs-local},
the graphs show that local reasoning clearly influences the total inference time.
In particular, at the growth of number of contexts, the
behaviour tends to be linear in the number of asserted triples.
%
%[It can be checked that the number of inferred triples is different from context to context]
While the data we have on \emph{ckr-owl-local}
are more limited, this behaviour seems to be confirmed by the trend lines.
On the other hand, OWL local reasoning seems to influence the reasoning time
with respect to the RDFS case: informally, this can be seen in the graph by the 
larger time overhead across points with a similar number of asserted triples
(i.e. on the same $x$ values) but a higher number of contexts.
%\lbreview{AGGIORNARE FINDINGS PER TS1}

%[We note that the study of the advantages (or not) to have different distribution
%of knowledge across contexts was not our goal in this test.]

%- - - - - - - - - - - - - - - - - - - - - - - - - - - - - - - - - - - - - - -
\smallskip\noindent
\textbf{TS2 and TS3: knowledge propagation evaluation.}
%
%\lbreview{AGGIORNA DESCRIZIONE E FINDINGS}
The second set of experiments we carried out
was aimed at answering question \textbf{RQ2}:
we wanted to establish the cost of knowledge propagation among contexts,
with respect to an increasing number of connections (i.e. $\eval$ expressions)
across contexts.
%
%\lbnote{Aggiorna con TS2!}
%To this aim, we generated 
%a second test set called \emph{TS2},
%shown in Table~\ref{tab:TS2}.
%
%\begin{table}[tp]
%\vspace{-2ex}
%\scriptsize\centering
%\begin{tabular}{|r|r|rrr|rrr|rrr|r|}
%\hline
%&& &&& \multicolumn{2}{l}{{\bf Global KB}} && \multicolumn{2}{l}{{\bf Local KBs}} && \\
%\hline
%{\bf Related} & {\bf Contexts}	&	{\bf Classes}	&	{\bf Roles}	&	{\bf Indiv.}	&	
%{\bf TBox}	&	{\bf RBox}	&	{\bf ABox}	&	
%{\bf TBox}	&	{\bf RBox}	&	{\bf ABox}	&	{\bf Total axioms}	\\
%\hline
%0 & 10	&	10	&	20	&	50	&	20	&	10	&	40	&	20	&	10	&	40	&	770	\\
%1 & 10	&	10	&	20	&	50	&	20	&	10	&	40	&	20	&	10	&	40	&	770	\\
%2 & 10	&	10	&	20	&	50	&	20	&	10	&	40	&	20	&	10	&	40	&	770	\\
%3 & 10	&	10	&	20	&	50	&	20	&	10	&	40	&	20	&	10	&	40	&	770	\\
%4 & 10	&	10	&	20	&	50	&	20	&	10	&	40	&	20	&	10	&	40	&	770	\\
%5 & 10	&	10	&	20	&	50	&	20	&	10	&	40	&	20	&	10	&	40	&	770	\\
%6 & 10	&	10	&	20	&	50	&	20	&	10	&	40	&	20	&	10	&	40	&	770	\\
%7 & 10	&	10	&	20	&	50	&	20	&	10	&	40	&	20	&	10	&	40	&	770	\\
%8 & 10	&	10	&	20	&	50	&	20	&	10	&	40	&	20	&	10	&	40	&	770	\\
%9 & 10	&	10	&	20	&	50	&	20	&	10	&	40	&	20	&	10	&	40	&	770	\\
%\hline
%\end{tabular}
%\vspace{2ex}
%\caption{Test set TS2.}
%\label{tab:TS2}
%\vspace{-5ex}
%\end{table}
%
To this aim, we generated two test sets, called \emph{TS2} and \emph{TS3}
structured as follows:
\begin{itemize}
\item 
  TS2 is composed by 100 CKRs, each of them with 100 contexts.
	Except for the triples needed for the definition of the contextual structure,
	both the global and local knowledge bases contain no randomly generated axioms.
	The CKRs inside TS2 are generated with an increasing number
	of contexts connections trough $\eval$ axioms (from no connections
	to the case of ``fully connected'' contexts).
  In particular, for $n = 100$ contexts and $k$ connections, 
	in each context $c_i$ we add axioms of the kind:
  %\begin{center}
    $$\eval(D_0, \{c_{i+1 (mod\ n)}\}) \subs D_1,\ \dots,\ \eval(D_0, \{c_{i+k (mod\ n)}\}) \subs D_1$$
  %\end{center}
  Moreover, in each context we add a fixed number of instances (10 in the case of TS2) of
  the local concept $D_0$, that will be propagated through contexts and added to 
  local $D_1$ concepts by the inference rules for the above $\eval$ expressions.
\item
  TS3 analogously contains 100 CKRs of 100 contexts and again no randomly generated
	global or local axioms. Differently from TS2, TS3 contains no $\eval$ axioms and
	the connections across contexts	are simulated by having multiple versions 
	of $D_0$ (namely $D_{0\-0}, \dots, D_{0\-99}$) to represent the local interpretation
	of the concept. Thus, for $n = 100$ contexts and $k$ connections, 
	in each context $c_i$ we add axioms of the kind $D_{0\-j} \subs D_1$	
	for $j \in \{i+1 (mod\ n), \dots, i+k (mod\ n)\}$.
	Also, not only we add to $c_i$ the 10 local instances of $D_{0\-i}$, but 
	we also ``pre-propagate'' instances of each $D_{0\-j}$ by explicitly
	adding them to the knowledge of $c_i$.	
\end{itemize}
We remark that the way of expressing ``contextualized symbols'' used in TS3
has been discussed and compared to the CKR representation in~\cite{BozGhiSer:KCAP2013}.

\begin{table}[th]
\vspace{-2ex}
\centering\scriptsize %\hspace{-9ex}
\begin{tabular}{|r|r@{\;\;\;}r@{\;\;\;}r@{\;\;\;}r|r@{\;\;\;}r@{\;\;\;}r@{\;\;\;}r|}
\hline
 & \multicolumn{2}{l}{{\bf TS2}} 
 &&& \multicolumn{2}{l}{{\bf TS3}} &&\\
\hline
{\bf Related}	&	
{\bf Triples}	&	{\bf Total}	&	{\bf Inf.}	&	{\bf Time} &	
{\bf Triples}	&	{\bf Total}	&	{\bf Inf.}	&	{\bf Time} \\
\hline
0	&	2803	&	3305	&	502	&	276	&	2803	&	3305	&	502	&	299	\\
4	&	4703	&	9205	&	4502	&	893	&	11703	&	16205	&	4502	&	577	\\
9	&	6703	&	16205	&	9502	&	1564	&	22703	&	32205	&	9502	&	1017	\\
14	&	8703	&	23205	&	14502	&	2245	&	33703	&	48205	&	14502	&	1450	\\
19	&	10703	&	30205	&	19502	&	2932	&	44703	&	64205	&	19502	&	1960	\\
24	&	12703	&	37205	&	24502	&	3467	&	55703	&	80205	&	24502	&	2580	\\
29	&	14703	&	44205	&	29502	&	4196	&	66703	&	96205	&	29502	&	3154	\\
34	&	16703	&	51205	&	34502	&	4847	&	77703	&	112205	&	34502	&	4099	\\
39	&	18703	&	58205	&	39502	&	5987	&	88703	&	128205	&	39502	&	4645	\\
44	&	20703	&	65205	&	44502	&	6223	&	99703	&	144205	&	44502	&	5488	\\
49	&	22703	&	72205	&	49502	&	6878	&	110703	&	160205	&	49502	&	6456	\\
54	&	24703	&	79205	&	54502	&	7689	&	121703	&	176205	&	54502	&	7545	\\
59	&	26703	&	86205	&	59502	&	8547	&	132703	&	192205	&	59502	&	8205	\\
64	&	28703	&	93205	&	64502	&	9076	&	143703	&	208205	&	64502	&	9159	\\
69	&	30703	&	100205	&	69502	&	9640	&	154703	&	224205	&	69502	&	10335	\\
74	&	32703	&	107205	&	74502	&	10711	&	165703	&	240205	&	74502	&	10992	\\
79	&	34703	&	114205	&	79502	&	11223	&	176703	&	256205	&	79502	&	11879	\\
84	&	36703	&	121205	&	84502	&	14611	&	187703	&	272205	&	84502	&	13088	\\
89	&	38703	&	128205	&	89502	&	12846	&	198703	&	288205	&	89502	&	13912	\\
94	&	40703	&	135205	&	94502	&	14999	&	209703	&	304205	&	94502	&	15064	\\
99	&	42703	&	142205	&	99502	&	14107	&	220703	&	320205	&	99502	&	15799	\\
\hline
\end{tabular}
\vspace{2ex}
\caption{Knowledge propagation results (extract) for test set TS2 and TS3.}
\label{tab:TS2results}
\vspace{-2ex}
\end{table}

\begin{figure}[t]\centering\vspace{-3ex}
\includegraphics[width=.824\columnwidth]{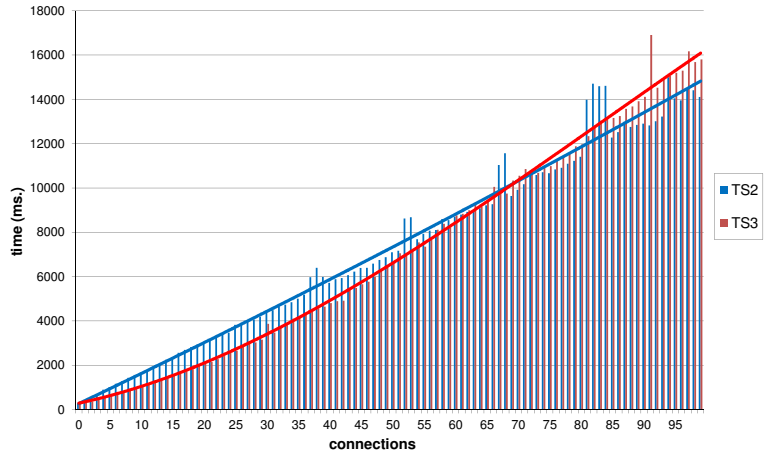}
\vspace{-2.5ex}
\caption{Knowledge propagation graphs for TS2 and TS3.}%
\label{fig:graph-TS2}%
\vspace{-3ex}
\end{figure}

We ran the CKR prototype for 5 independent runs on TS2 and TS3,
only considering \emph{ckr-owl-local} ruleset.
An extract of the results of experiments
on the two test sets is reported in Table~\ref{tab:TS2results}:
CKRs in the two sets are ordered with respect to the number of
relations across contexts; for each CKR, the numbers of asserted,
total and inferred triples are shown, followed by the (average) closure time in milliseconds.
To facilitate the analysis of the results, we plotted such data
in histograms in Figure~\ref{fig:graph-TS2}. 
The $x$ axis represents the number of local connections, while the
$y$ axis shows the time in milliseconds.
Again, to better visualize the behaviour of the series, we plotted a trend line
for each of the series, calculated by polynomial 
%\lbreview{AGGIORNARE GRAFICI PER TS2}
%\lbreview{AGGIORNARE $R^2$ e \% PER TS2}
regression\footnote{Average $R^2$ value across the two approximations
is $\geq 0,989$.}.

%
%\noindent
%A number of conclusions can be derived from the presented results and graphs.
From the graph of TS2, we can note that knowledge propagation cost 
depends linearly on the number of connections: from the data in Table~\ref{tab:TS2results}
 we can calculate that the average increase in closure time
for $k$ local connections (for each context) w.r.t. the base case of 0 connections
amounts to $(51,\!2 \cdot k) \%$.
The comparison with TS3 %and the two trend lines
confirms the compactness of a contextualized representation of symbols 
(cfr. findings in~\cite{BozGhiSer:KCAP2013}): 
in fact, note that for an equal number of connections, 
the number of inferences in both TS2 and TS3 cases is equal,
but TS3 always require a larger number of asserted triples.
%and generally (in the $73,\!7\%$ of cases) require more inference time.
Also, the graph clearly shows that %generally (in $86\%$ of cases) 
TS3 grows more than linearly:
for a small number of connections the knowledge propagation in TS2 requires 
more inference time ($14,\!9\%$ more, on average),
but with the growth of local connections (at $\sim\!68\%$ of number of contexts) the 
cost of TS3 local reasoning surpasses the propagation overhead.

%\lbnote{CONTROLLARE ANDAMENTO LINEARE DELLA TS2?}

%\lbreview{AGGIORNARE FINDINGS PER TS2}

%###################################
\vspace{-.7ex}

%-----------------------------------------------------------------------------

%  \input{sec/conclusion}

%-----------------------------------------------------------------------------
\section{Conclusions and Future Works}
\label{sec:conclusion}

%\vspace{-1ex}

% # Summary #
In this paper we provided a first evaluation for the performance of
the RDF based implementation of the CKR framework.
In first experiment we evaluated the scalability of the current version of the
prototype under different reasoning regimes.
The second experiment was aimed at evaluating the cost of intra-context knowledge propagation
and its relation to its simulation by ``reification'' of contextualized symbols.

% # Future works #
Some further experimental evaluations can be interesting to be carried out
over our contextual model: one of these can regard the study of the cost and advantages 
of the separation of the same amount of knowledge across different contexts.
With respect to the current CKR implementation, 
the scalability experiments clearly showed that the current naive strategy 
(defined by a direct translation of the formal calculus) might 
not be suitable for a real application of the full reasoning to large scale datasets.
In this regard, we are going to study different evaluation strategies and 
optimizations to the current strategy and evaluate the results with respect to 
the naive case. One of such possible optimizations can regard a ``pay-as-you-go'' strategy,
in which inference rules are activated only for constructs that are recognized in the 
local language of a context.

\vspace{-2ex}
  
%-----------------------------------------------------------------------------
% Acknowledgements

%\medskip\noindent
%{\bf Acknowledgments.}
%The research leading to these results has received funding from the 
%European Union's Seventh Framework Programme (FP7/2007-2013) under grant 
%agreement no.257641 (PlanetData NoE).

%-----------------------------------------------------------------------------

%\nocite{BozzatoEiterSerafini:13}
\bibliographystyle{splncs03}
\bibliography{bibliography}

%-----------------------------------------------------------------------------
%\newpage
%\appendix
%\input{sec/appendix}

\end{document}